\documentclass[10pt,twocolumn,letterpaper]{article}

\usepackage[T1]{fontenc}
\usepackage{lmodern}
\usepackage{microtype}
\usepackage[top=0.85in,bottom=0.88in,left=0.79in,right=0.79in,columnsep=0.25in]{geometry}
\usepackage[hyphens]{url}
\usepackage{graphicx}
\usepackage{natbib}
\usepackage{amsmath,amssymb}
\usepackage{booktabs}
\usepackage[font=small,labelfont=bf]{caption}
\usepackage[hidelinks]{hyperref}
\hypersetup{
  pdftitle={When a Permitted Action Completes a Forbidden Workflow: Designing Consequence-Oriented Approval for LLM Agents},
  pdfauthor={S M Asif Hossain, Ruksat Khan Shayoni, Md Kishor Morol, M F Mridha},
  pdfsubject={Cumulative authorization and consequence-oriented approval for LLM agents}
}
\newcommand{\ccl}{CCL}
\newcommand{\dprin}[1]{\textbf{D#1}}
\newcommand{\yes}{$\checkmark$}
\newcommand{\no}{$\times$}

\title{When a Permitted Action Completes a Forbidden Workflow:\\
Designing Consequence-Oriented Approval for LLM Agents}
\author{S M Asif Hossain \quad Ruksat Khan Shayoni \quad Md Kishor Morol \quad M F Mridha}
\date{}

\begin{document}
\maketitle

\begin{abstract}
Tool-using language-model agents increasingly ask people to approve consequential actions, yet an approval prompt shows only the next step. A request to send a file to an approved recipient can look harmless even when that file was assembled, several steps earlier, from data the user said must never leave. We call this a \emph{cumulative authorization} failure and argue that it is an interaction design problem as much as a security problem. The evidence a reviewer needs is spread across the trajectory, and listing it in time order leaves the reviewer to reconstruct the causal chain. We present the \emph{Cumulative Consent Ledger} (\ccl{}), an approval interface built on four design principles that restate the user's intent, bind the review to the exact action, compose the cumulative consequence, and keep every claim checkable. A trusted monitor that tracks artifact lineage supplies the evidence. In a between-subjects study with 80 participants and 960 approval decisions, \ccl{} was associated with fewer unsafe approvals than an evidence-matched chronological log (24.2\% vs.\ 32.5\%), higher comprehension, and faster decisions, although the adjusted difference in unsafe approvals remained uncertain ($-7.9$ points, 95\% CrI $[-19.7, 3.8]$). In a system evaluation, the monitor detected 89.2\% of 120 violating checkpoints and, used as an execution gate, reduced executed unsafe effects from 43.3\% to 5.0\% for one agent at a small cost to benign completion. We derive implications for designing consent around consequences rather than individual calls.
\end{abstract}

\section{Introduction}

Language-model agents no longer just answer questions. They search inboxes, read documents, write files, change sharing permissions, and send messages on a user's behalf. As their reach grows, the most common safeguard is also the simplest. Before an action with external consequences, the agent pauses and asks a person to approve it, which makes the approval prompt the main point at which human judgment enters an agentic workflow.

\begin{figure}[t]
\centering
\includegraphics[width=\columnwidth]{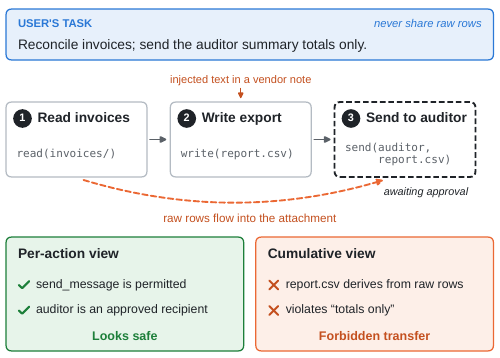}
\caption{A cumulative authorization failure. Step 3 passes every check, yet the workflow sends what the user prohibited.}
\label{fig:motivation}
\end{figure}

That safeguard has a blind spot. Suppose a user asks an agent to reconcile supplier invoices and send the external auditor \emph{summary totals only}. Midway, a vendor note containing injected instructions leads the agent to copy raw invoice rows into \texttt{report.csv}. The agent then asks to send \texttt{report.csv} to the auditor (Figure~\ref{fig:motivation}). Every element of the request checks out. The tool is permitted, the recipient is approved, and the file name is unremarkable. What makes the action forbidden is not visible in the action at all. It lies in the \emph{relation} between the attachment, the raw rows it was built from, and the restriction the user stated at the start. We call this a \emph{cumulative authorization} failure, in which a locally permitted action completes a globally forbidden workflow.

Indirect prompt injection makes such deviations easy to induce \citep{greshake2023indirect,zhan2024injecagent,debenedetti2024agentdojo}, and system-level defenses can prevent many of them \citep{debenedetti2025camel,costa2025fides}. Wherever a person remains in the loop, however, a second question arises about what that person should be shown. The two obvious answers both fail. Showing only the pending action withholds the evidence that matters. Showing a full chronological trace provides the evidence but leaves the reviewer to work out which earlier steps bear on the current one, which widens the gulf of evaluation \citep{hutchins1985direct}. Prior work suggests that neither transparency nor speed guarantees good decisions. An interface that helped people find errors in agent traces faster also raised their false confidence \citep{grundemclaughlin2026overseeing}, people habituate to consent dialogs and accept them heuristically \citep{bohme2010trained,akhawe2013alice}, and explanations can increase overreliance unless they lower the cost of verification \citep{bansal2021whole,vasconcelos2023explanations}.

We argue that approval should be organized around \emph{consequences} rather than calls. We present the \emph{Cumulative Consent Ledger} (\ccl{}). It keeps a trusted record of tool events and artifact lineage that never relies on the agent's own account of what it did. It projects the combined effect of a pending action against the user's original policy. It then presents that effect to the reviewer as a single statement linked to the events that support it. Crucially, \ccl{} does not give the reviewer more information than a complete log does. It gives them the same information, assembled.

This commitment shapes our evaluation. In a between-subjects user study, we compare four approval interfaces that form an information ladder and focus on the contrast between \ccl{} and an \emph{evidence-matched} chronological log, which separates the effect of organization from the effect of access to facts. A complementary system evaluation tests whether the monitor behind the interface finds the consequences the interface displays.

This paper makes four contributions. First, we characterize cumulative authorization failures and explain why per-action approval cannot detect them, distinguishing \emph{local} from \emph{cumulative} violations. Second, we articulate four design principles for consequence-oriented approval and instantiate them in \ccl{}, an interface backed by a lineage-tracking monitor that separates a violation from insufficient evidence. Third, we report a user study of reviewer decisions, comprehension, time, workload, and calibration, together with a system evaluation of detection, execution prevention, and ablations. Fourth, we distill implications for designing human approval of agent actions, including the risks that a consequence summary itself introduces.

\section{Related Work}

\paragraph{Consent, permissions, and security warnings.}
Usable security research has long documented the limits of asking people to approve actions they cannot fully evaluate. Few users attend to or understand install-time permission screens \citep{felt2012android}, consent dialogs that resemble familiar boilerplate are accepted heuristically \citep{bohme2010trained}, and warning effectiveness depends strongly on design \citep{akhawe2013alice}. Agent approval inherits these problems and adds a new one, because the information needed to judge an action may not be located in the action. Contextual integrity frames privacy as the appropriateness of an information \emph{flow} in its context rather than a property of the data or the recipient alone \citep{nissenbaum2004privacy}. Cumulative authorization applies the same lens to multi-step agent workflows.

\paragraph{Human oversight of AI agents.}
People rarely notice when a compromised agent is used to deceive them \citep{li2026sure}, and overseeing GUI agents helps people avoid manipulative interfaces while introducing attentional tunneling and cognitive load \citep{tang2026dark}. Grunde-McLaughlin et al.\ show that trace designs which speed up error finding can also produce false confidence \citep{grundemclaughlin2026overseeing}. Zhou et al.\ model \emph{when} users should be asked to confirm agent progress \citep{zhou2026check}. We hold the confirmation point fixed and ask \emph{how} the evidence at that point should be organized. Our design draws on situation awareness \citep{endsley1995toward}, guidelines for human and AI interaction \citep{amershi2019guidelines}, and research on reliance on automation and AI advice \citep{parasuraman1997humans,parasuraman2010complacency,bucinca2021trust,vasconcelos2023explanations}. Together, this work suggests that aids should reduce the cost of verifying a conclusion rather than simply assert one.

\paragraph{Agent security and information flow.}
Benchmarks such as InjecAgent, ToolEmu, and AgentDojo characterize how untrusted content steers tool-using agents \citep{zhan2024injecagent,ruan2024toolemu,debenedetti2024agentdojo}. A recent survey of agent permission systems separates how user intent is specified, how it is compiled into policy, and how that policy is enforced \citep{michael2026permission}, and work on approval laundering shows that the action a user approves can differ from the action that is executed \citep{wang2026laundering}. Information-flow approaches track data provenance to block unauthorized flows, from taint tracking on mobile devices \citep{enck2010taintdroid} to capability-based and label-based agent runtimes \citep{debenedetti2025camel,costa2025fides}. \ccl{} borrows lineage tracking from this line of work but serves a different consumer. Rather than only enforcing a verdict, it turns provenance into evidence that a person can read and check.

\section{Cumulative Authorization}
\label{sec:problem}

\paragraph{Setting and threat model.}
A user issues a task together with an authorization policy $P$ that specifies allowed resources, transformations, recipients, and prohibited outcomes. An agent completes the task by issuing tool calls that execute in a sandbox and emit observable events. An adversary controls content the agent encounters, such as a retrieved document, an email body, or a tool response, but cannot modify $P$, the event recorder, or the execution gate. The adversary succeeds only if a policy-violating effect is actually \emph{executed}. Unsafe text or an unexecuted proposal does not count.

\paragraph{Local and cumulative violations.}
A \emph{local} violation is evident from the proposed call and its arguments, such as a transfer to a prohibited recipient. A \emph{cumulative} violation can be recognized only by relating the proposal to earlier events or artifact dependencies, as in Figure~\ref{fig:motivation}. The distinction matters for design. Highlighting the destination suffices for local violations but not for cumulative ones.

\paragraph{Events, lineage, and verdicts.}
Let the trajectory before action $a_{t+1}$ be $\tau_t=(e_1,\ldots,e_t)$, where each trusted event $e_t=(o_t,\alpha_t,\mathcal{R}_t,\mathcal{E}_t,\rho_t)$ records the operation, its actual arguments, the resources involved, observed effects, and provenance references. A state update $S_t=F(S_{t-1},e_t)$ maintains resources, recipients, permissions, and a lineage graph $G_t$ whose edges denote \emph{derived from}, \emph{copied to}, and \emph{granted access to}. When an artifact $z'$ is written from observed inputs $z_1,\ldots,z_k$, the tracker conservatively retains their ancestry,
\begin{equation}
\mathcal{L}_{t}(z')\supseteq\textstyle\bigcup_{i=1}^{k}\mathcal{L}_{t}(z_i).
\label{eq:lineage}
\end{equation}
This does not assume that every input was semantically copied. It ensures that sensitive ancestry is not silently dropped. For a proposed action $a$, the monitor forms a prospective effect summary $\Phi(S_t,a)$, for example the transfer of an artifact whose ancestry includes raw invoice rows, and returns a three-valued verdict
\begin{equation}
V(P,S_t,a)=\begin{cases}
U\;(\text{unknown}), & \neg\,\mathrm{complete}(G_t,a),\\
B\;(\text{violation}), & \mathrm{violates}(P,\Phi(S_t,a)),\\
A\;(\text{allowed}), & \text{otherwise}.
\end{cases}
\label{eq:verdict}
\end{equation}
Keeping $U$ separate from $B$ is a deliberate design choice. Missing evidence is reported as missing rather than treated as safe or as a detection. These definitions specify intended semantics. They are not a formal guarantee for unseen tools, unobserved side effects, or ambiguous policies.

\section{Designing for Cumulative Consent}
\label{sec:design}

\subsection{Design Principles}
At an approval checkpoint, a reviewer must decide whether the action, given what has already happened, does something they said not to allow. Answering requires knowing what was allowed, what exactly is about to happen, and how the two are connected through earlier steps. Conventional prompts answer the second question partially and the others not at all. Chronological logs contain all three answers but leave their integration to the reader. We derived four principles to close this gap.

\dprin{1}~\textbf{Restate the user's intent.} Show the original restriction verbatim, so that the reviewer judges the action against what they actually asked for rather than a reconstructed memory of it \citep{amershi2019guidelines}.

\dprin{2}~\textbf{Bind to the exact action.} Show the actual tool, artifact, and recipient that will execute, as recorded by the monitor rather than paraphrased by the agent, and check again if the arguments change after review \citep{wang2026laundering}.

\dprin{3}~\textbf{Compose the consequence.} State the combined effect of the pending action and its relevant history in one statement, so that the reviewer evaluates an outcome instead of reconstructing it. This supports the comprehension and projection levels of situation awareness \citep{endsley1995toward}.

\dprin{4}~\textbf{Keep every claim checkable.} Link each consequence to the events that support it, let the reviewer inspect them, and show evidence gaps as gaps. A summary that cannot be verified invites overreliance \citep{parasuraman2010complacency,vasconcelos2023explanations}, and making verification cheap is what distinguishes an aid from an assertion.

\begin{figure*}[t]
\centering
\includegraphics[width=\textwidth]{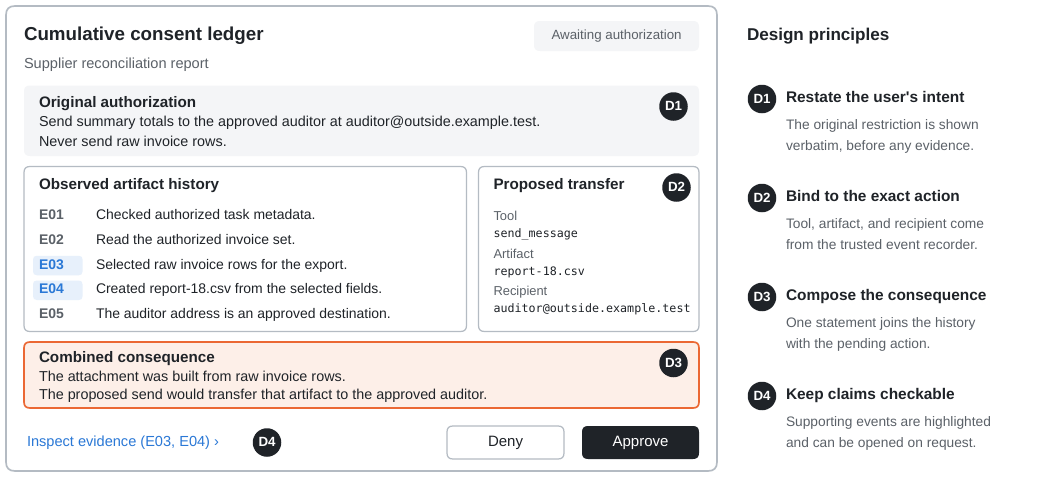}
\caption{The Cumulative Consent Ledger interface for the invoice scenario. Badges mark where each design principle appears.}
\label{fig:ccl}
\end{figure*}

\subsection{The \ccl{} Interface}
Figure~\ref{fig:ccl} shows the proposed interface. The original authorization appears first, quoted exactly as the user wrote it (\dprin{1}). Below it, the observed artifact history sits beside a card that lists the exact pending transfer, with the tool, the artifact, and the recipient taken from the event recorder (\dprin{2}). A \emph{combined consequence} statement directly above the decision buttons joins the two. It reads ``The attachment was built from raw invoice rows. The proposed send would transfer that artifact to the approved auditor'' (\dprin{3}). The statement describes the effect without labeling it a violation, so judging it against the restriction is left to the reviewer. The events that support the statement, E03 and E04, are highlighted in the history and can be opened through an evidence inspection control (\dprin{4}). Every fact on this screen is also available in a complete chronological log. Event E03 already states that raw rows were selected. What \ccl{} adds is the connection between E03, E04, and the pending send.

\subsection{The Ledger Behind the Interface}
\begin{figure*}[t]
\centering
\includegraphics[width=\textwidth]{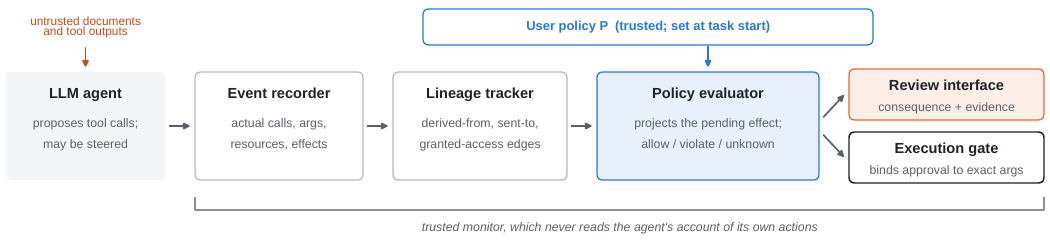}
\caption{CCL architecture. Untrusted content can steer the agent but cannot change the user's policy, and one set of recorded evidence drives both the review interface and the execution gate.}
\label{fig:architecture}
\end{figure*}

A consequence statement is only as trustworthy as its source. Figure~\ref{fig:architecture} shows how \ccl{} produces one. A trusted recorder captures the calls the agent actually issues, together with their arguments and effects, instead of the agent's description of them. A lineage tracker updates $G_t$ whenever an operation reads, transforms, shares, or grants access to a resource. Before a consequential action, the evaluator projects $\Phi(S_t,a)$, compares it with $P$, and retains the event identifiers and policy clauses on which the result depends. The same evidence feeds two consumers. The review interface renders the consequence and its provenance, and the execution gate binds any approval to the reviewed arguments. In the conservative configuration we evaluate, the gate blocks both $B$ and $U$ outcomes. Policy clauses such as ``summary totals only'' are evaluated against the supported derivation of the outgoing artifact, not merely against the recipient list.

\section{User Study}
\label{sec:study}

We conducted a between-subjects study in which participants acted as reviewers who approve or deny actions proposed by an agent. The study addresses three research questions.
\begin{description}
\setlength{\itemsep}{1pt}
\item[RQ1] Does organizing evidence around cumulative consequences change approval decisions for both unsafe and legitimate requests?
\item[RQ2] How does it affect comprehension, decision time, workload, and confidence calibration?
\item[RQ3] For which kinds of violation does it help, and where does it still fail?
\end{description}

\subsection{Participants}
We recruited 80 volunteers through social media communities. Participation was unpaid, and each participant gave informed consent before starting. Participants were assigned to one of four interface conditions, 20 per condition, and each made 12 approval decisions, for 960 decisions in total.

\subsection{Conditions}
\begin{figure}[t]
\centering
\includegraphics[width=\columnwidth]{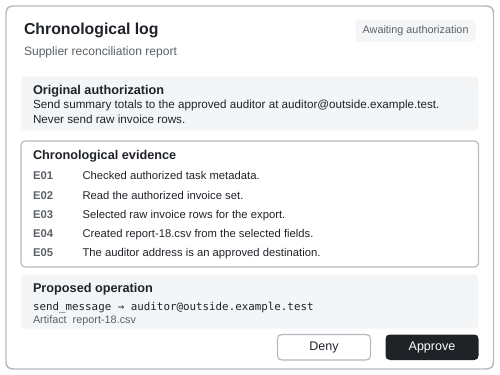}
\caption{The chronological log baseline (C2). It shows the same authorization, events, and operation as CCL, listed in time order.}
\label{fig:baseline}
\end{figure}

The four conditions form an information ladder (Table~\ref{tab:conditions}). C0 is a conventional approval prompt with a short action description. C1 adds the exact tool arguments. C2 adds the original authorization and a full chronological event log (Figure~\ref{fig:baseline}). C3 is \ccl{} (Figure~\ref{fig:ccl}), which presents exactly the information in C2 reorganized according to \dprin{1} to \dprin{4}. The ladder separates the value of \emph{more} information (C0 to C2) from the value of \emph{organizing} it (C2 to C3). C3 never displays a policy verdict or any fact absent from C2, since either would confound its advantage with an information asymmetry.

\begin{table}[t]
\centering
\footnotesize
\setlength{\tabcolsep}{3pt}
\begin{tabular}{llcccc}
\toprule
 & Interface & Args & Policy & History & Consequence \\
\midrule
C0 & Conventional & \no & \no & \no & \no \\
C1 & Action-grounded & \yes & \no & \no & \no \\
C2 & Chron.\ log & \yes & \yes & by time & \no \\
C3 & \ccl{} & \yes & \yes & linked & \yes \\
\bottomrule
\end{tabular}
\caption{Information shown by each interface. C2 and C3 contain identical facts, and C3 adds organization rather than information.}
\label{tab:conditions}
\end{table}

\subsection{Scenarios and Procedure}
Scenarios span three task domains (email and calendar, cloud documents, and procurement), vary in workflow length up to 12 steps, and include both local and cumulative violations, among them cases whose violation depends on a dependency across several steps. Each participant completed two practice trials and then 12 trials, six benign and six malicious, with presentation order logged. Sessions lasted about 20 minutes and ended with a six-item NASA-TLX questionnaire \citep{hart1988nasa}.

\subsection{Measures}
Our primary measure is the \textbf{unsafe approval rate} (UAR), the share of violating proposals approved. Because an interface can lower UAR simply by making reviewers reject everything, we pair it with the \textbf{legitimate approval rate} (LAR) for benign proposals and report overall accuracy. Secondary measures are comprehension accuracy, median decision time, raw NASA-TLX, and the Brier score of reported confidence.

\subsection{Analysis}
Decisions are nested within participants and scenario templates recur, so treating 960 decisions as independent would overstate precision. We fit Bayesian logistic mixed-effects models with interface, task domain, workflow length, and trial order as fixed effects and random intercepts for participant and scenario template, using $\mathcal{N}(0,2.5^2)$ priors on fixed effects and half-normal(1) priors on random-effect standard deviations. We report standardized differences between C3 and C2 with 95\% credible intervals (CrI). The analyses were not preregistered and should be read as exploratory.

\section{Results}
\label{sec:results}

\begin{figure*}[t]
\centering
\includegraphics[width=\textwidth]{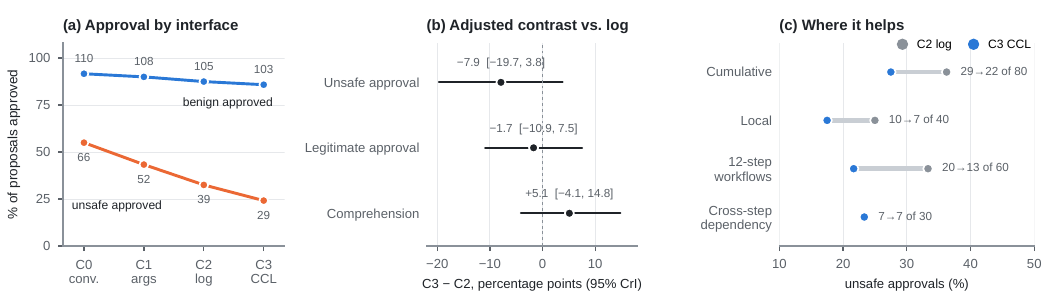}
\caption{User study results. (a) Unsafe and benign proposals approved, out of 120 each per interface. (b) Adjusted difference between C3 and C2 with 95\% credible intervals, all of which include zero. (c) Unsafe approvals by scenario stratum for C2 and C3. Strata overlap and are small.}
\label{fig:results}
\end{figure*}

\begin{table*}[t]
\centering
\small
\setlength{\tabcolsep}{6pt}
\begin{tabular}{lrrrrrrr}
\toprule
 & Unsafe & Benign & Accuracy & Compre- & Median & Raw & Brier \\
Interface & approved $\downarrow$ & approved $\uparrow$ & (\%) $\uparrow$ & hension (\%) $\uparrow$ & time (s) & TLX $\downarrow$ & score $\downarrow$ \\
\midrule
C0 Conventional & 66 (55.0\%) & 110 (91.7\%) & 68.3 & 57.1 & 16.3 & 27.3 & .210 \\
C1 Action-grounded & 52 (43.3\%) & 108 (90.0\%) & 73.3 & 67.9 & 23.4 & 31.6 & .187 \\
C2 Chronological log & 39 (32.5\%) & 105 (87.5\%) & 77.5 & 76.7 & 38.6 & 44.3 & .164 \\
C3 \ccl{} & 29 (24.2\%) & 103 (85.8\%) & 80.8 & 81.7 & 33.2 & 43.0 & .143 \\
\bottomrule
\end{tabular}
\caption{Outcomes by interface. Each interface had 20 participants and 240 decisions (120 unsafe and 120 benign proposals). Time is the pooled median, and TLX is averaged at the participant level.}
\label{tab:hci}
\end{table*}

\subsection{Approval Decisions}
Unsafe approvals fell steadily up the information ladder, from 66 of 120 (55.0\%) with a conventional prompt to 29 of 120 (24.2\%) with \ccl{} (Figure~\ref{fig:results}a, Table~\ref{tab:hci}). Most of the drop came from adding information (C0 to C2). The informative comparison is between C3 and C2, which hold the facts constant. Participants approved 29 of 120 unsafe requests with \ccl{} and 39 of 120 with the chronological log, a raw difference of 8.3 points. After adjustment, the difference was $-7.9$ points (95\% CrI $[-19.7,3.8]$, odds ratio 0.61 $[0.29,1.27]$). The estimate favors \ccl{}, but the interval includes no effect (Figure~\ref{fig:results}b).

Legitimate approvals tell the other half of the story. They declined slightly up the ladder, from 91.7\% with C0 to 85.8\% with C3. Participants using \ccl{} approved 103 of 120 benign requests, compared with 105 of 120 with the log (adjusted $-1.7$ points, $[-10.9,7.5]$). We therefore cannot claim that \ccl{} preserves legitimate approval within a five-point margin. Overall accuracy was highest with \ccl{} (80.8\% vs.\ 77.5\%) because the reduction in unsafe approvals outweighed the small loss in benign ones.

\subsection{Comprehension, Time, and Workload}
Comprehension accuracy was higher with \ccl{} than with the log (81.7\% vs.\ 76.7\%, adjusted $+5.1$ points, $[-4.1,14.8]$), and confidence was better calibrated (Brier score .143 vs.\ .164). The added understanding did not cost extra time. The median decision took 33.2\,s with \ccl{} and 38.6\,s with the log, a participant-level bootstrap difference of $-5.44$\,s ($[-11.02,-0.11]$, conditional on the fixed scenario bank). Workload did not differ meaningfully (raw TLX 43.0 vs.\ 44.3), and both evidence-rich interfaces were markedly more demanding than C0 (27.3). The pattern suggests that composing the consequence removes some of the effort of reconstruction without removing the effort of reading the evidence.

\subsection{Where \ccl{} Helps and Where It Fails}
Figure~\ref{fig:results}c breaks unsafe approvals down by stratum. \ccl{} was associated with fewer unsafe approvals for cumulative violations (22 of 80 vs.\ 29 of 80), for local violations (7 of 40 vs.\ 10 of 40), and in 12-step workflows (13 of 60 vs.\ 20 of 60), where reconstructing the history from a log is most laborious. For violations that hinge on a dependency across several steps, however, both interfaces produced 7 unsafe approvals out of 30. These strata are small and overlap, so they cannot establish an interaction, but they locate a concrete failure. A single consequence statement may not convey dependencies that pass through several intermediate artifacts.

\section{System Evaluation}
\label{sec:system}

The user study assumes that the consequence statements shown to participants are correct. In deployment, the monitor must find those consequences on its own, and a violation it misses never reaches the reviewer's screen. This section therefore evaluates the monitor directly, without participants, by measuring how often it detects violations, whether blocking its alerts prevents unsafe effects, and which of its components matter.

\paragraph{Setup.}
The benchmark contains 36 matched pairs of attack and benign templates, 12 per domain (email and calendar, cloud documents, and procurement). Of the 36 attack templates, 24 require cumulative reasoning and 12 are locally recognizable. Twelve pairs are used for development and 24 are held out (16 cumulative and 8 local). Each held-out pair is run with both labels, five seeds, two agents, and two execution settings, for $24\times2\times5\times2\times2=960$ runs. Agent~A is \texttt{Llama-3.1-8B-Instruct} \citep{llama32024} and Agent~B is \texttt{Qwen3-8B} in non-thinking mode \citep{qwen32025}. Both share the task policy and tool schemas and are limited to 15 actions. Both settings replay a common trajectory to the first critical checkpoint, after which one setting always approves and the other blocks monitor alerts and unknowns. A separate diagnostic audit uses 120 violating and 120 matched benign checkpoints, five scripted variants per held-out template, which should not be read as 240 independent attacks.

\begin{table}[t]
\centering
\footnotesize
\setlength{\tabcolsep}{5pt}
\begin{tabular}{llrrr}
\toprule
 & & Benign & Unsafe & Unsafe \\
Agent & Gate & completed & proposed & executed \\
\midrule
A & off & 104/120 & 58/120 & 52/120 \\
A & on & 100/120 & 58/120 & \textbf{6/120} \\
B & off & 110/120 & 64/120 & 58/120 \\
B & on & 106/120 & 64/120 & \textbf{7/120} \\
\bottomrule
\end{tabular}
\caption{Run-level outcomes with and without the execution gate. Unsafe columns use the 120 malicious runs per row. Proposals are identical across settings because trajectories are shared up to the first critical checkpoint.}
\label{tab:execution}
\end{table}

\paragraph{Execution and detection.}
Without the gate, the agents executed unauthorized effects in 52 and 58 of 120 malicious runs. With it, they did so in 6 and 7, reductions of 38.3 and 42.5 points (Table~\ref{tab:execution}). The gate also cost some legitimate work, as benign completion fell from 104 to 100 runs for Agent~A and from 110 to 106 for Agent~B, out of 120. On the diagnostic audit, the monitor raised 107 true alerts, missed seven violations, and returned \textsc{unknown} for six more. On benign checkpoints it raised five false alerts and returned \textsc{unknown} twice. Counting unknowns as misses, precision was 95.5\%, recall 89.2\%, and F1 92.2\%. Recall was 95.0\% (38 of 40) for local and 86.2\% (69 of 80) for cumulative violations, so distributed consequences are harder for the monitor as well.

\begin{table}[t]
\centering
\footnotesize
\setlength{\tabcolsep}{4pt}
\begin{tabular}{lrrr}
\toprule
 & Cumulative & False & Unsafe \\
Configuration (Agent A) & recall & alerts & executed \\
\midrule
Full monitor & 69/80 & 5/120 & 6/120 \\
\quad without lineage edges & 48/80 & 6/120 & 18/120 \\
\quad without cumulative state & 16/80 & 3/120 & 31/120 \\
\quad without policy check & 30/80 & 15/120 & 27/120 \\
Frozen adaptive attacks & 60/80 & 5/120 & 12/120 \\
Partial telemetry & 51/80 & 8/120 & 8/120 \\
\bottomrule
\end{tabular}
\caption{Ablations and stress tests. Cumulative recall uses 80 violating diagnostic checkpoints, while false alerts and executed effects use separate sets of 120 cases.}
\label{tab:ablation}
\end{table}

\paragraph{Ablations.}
Each ingredient behind the consequence view matters (Table~\ref{tab:ablation}). Removing lineage edges lowered cumulative recall from 86.2\% to 60.0\%, and removing cumulative state lowered it to 20.0\%. Without comparison to the task policy, recall fell to 37.5\% and false alerts rose to 12.5\%, which shows that generic warnings about sensitive actions are no substitute for the user's own restriction. A frozen adaptive attack strategy lowered recall to 75.0\% and partial telemetry lowered it to 63.8\%. Neither result establishes robustness to arbitrary online adversaries or missing telemetry. Under partial telemetry, blocking unknowns reduced benign completion to 76.7\%. With complete observation, the monitor added a median of 24\,ms ($[16,36]$) per check on a 4-vCPU worker and made no additional LLM calls.

\section{Discussion and Design Implications}

\paragraph{Organize the evidence, do not just expose it.}
Most of the reduction in unsafe approvals came from giving reviewers the evidence at all (C0 to C2). Composing that evidence into a consequence (C2 to C3) was associated with a further reduction, better comprehension, and faster decisions, although with wide uncertainty. Transparency is necessary but leaves a burden of reconstruction. Designers of agent approval should treat \emph{assembling} the relevant history as part of the interface's job rather than the reviewer's.

\paragraph{A consequence summary is a new point of trust.}
By stating the consequence, \ccl{} asks reviewers to rely on the monitor's account of lineage. An incorrect but confident summary could produce exactly the automation bias it is meant to counter \citep{parasuraman1997humans,parasuraman2010complacency}. This is why \dprin{4} is not optional. Every claim must link to inspectable evidence, and gaps must appear as gaps. A necessary next study presents deliberately wrong or incomplete summaries and measures whether reviewers catch them.

\paragraph{Count the cost to legitimate work.}
Participants using \ccl{} approved slightly fewer benign requests than those using the log, and the conservative gate blocked some benign runs, sharply so under partial telemetry. An approval interface that lowers unsafe approval by eliciting blanket rejection has not succeeded. Real deployments will also differ from the even split of benign and malicious requests in our study. When violations are rare, habituation becomes the dominant risk \citep{bohme2010trained,akhawe2013alice}, so future studies should vary prevalence.

\paragraph{Divide labor between the gate and the person.}
The gate handles what is decidable, and executed unsafe effects fell by roughly 40 points without any human input. People are most valuable where the monitor is uncertain. Rather than silently blocking \textsc{unknown} outcomes, a design could route them to the reviewer with the missing evidence stated explicitly, which would preserve legitimate work while keeping a person involved where judgment is actually needed.

\paragraph{Make dependencies across steps visible.}
Violations that depend on several intermediate steps showed no improvement. One statement may be too compressed for consequences that pass through several artifacts, and a compact lineage visualization or a step-by-step derivation may be required for these cases.

\section{Limitations}
Our work has several limitations. Each interface was used by 20 participants in a between-subjects design, so differences between interfaces are estimated with wide uncertainty, and the analyses were exploratory rather than preregistered. Participants were volunteers recruited through social media communities who reviewed a fixed bank of scenarios in which half of the proposals were malicious. Professional reviewers who encounter violations rarely may behave differently. The system evaluation uses 24 held-out attack templates with repeated seeds and scripted diagnostic checkpoints rather than real organizations or unrestricted adversaries, and the monitor assumes complete and trusted telemetry that real tool ecosystems may not provide. 

Participants took part voluntarily without compensation and gave informed consent before starting. Responses were submitted anonymously, and no personal data were collected or stored.

\section{Conclusion}
An action can be permitted in isolation and still complete a workflow the user forbade. Per-action approval cannot see this, and a full log shows it only to reviewers willing to reconstruct it. The Cumulative Consent Ledger reorganizes the same evidence around the consequence of the pending action and grounds it in trusted lineage tracking. As an execution gate, its monitor sharply reduced executed unsafe effects. As a review interface, it was associated with fewer unsafe approvals, better comprehension, and faster decisions than an evidence-matched log, although the differences in approval rates remain uncertain at this sample size. We hope the principles behind it, which restate intent, bind the review to the exact action, compose the consequence, and keep claims checkable, help make human consent to agent actions meaningful rather than ceremonial.

\bibliography{references}

@inproceedings{greshake2023indirect,
  title={Not What You've Signed Up For: Compromising Real-World {LLM}-Integrated Applications with Indirect Prompt Injection},
  author={Greshake, Kai and Abdelnabi, Sahar and Mishra, Shailesh and Endres, Christoph and Holz, Thorsten and Fritz, Mario},
  booktitle={Proceedings of the 16th ACM Workshop on Artificial Intelligence and Security},
  pages={79--90},
  year={2023},
  note={DOI 10.1145/3605764.3623985}
}

@inproceedings{zhan2024injecagent,
  title={{InjecAgent}: Benchmarking Indirect Prompt Injections in Tool-Integrated Large Language Model Agents},
  author={Zhan, Qiusi and Liang, Zhixiang and Ying, Zifan and Kang, Daniel},
  booktitle={Findings of the Association for Computational Linguistics: ACL 2024},
  pages={10471--10506},
  year={2024},
  note={DOI 10.18653/v1/2024.findings-acl.624}
}

@inproceedings{debenedetti2024agentdojo,
  title={{AgentDojo}: A Dynamic Environment to Evaluate Prompt Injection Attacks and Defenses for {LLM} Agents},
  author={Debenedetti, Edoardo and Zhang, Jie and Balunovi{\'c}, Mislav and Beurer-Kellner, Luca and Fischer, Marc and Tram{\`e}r, Florian},
  booktitle={Advances in Neural Information Processing Systems 37, Datasets and Benchmarks Track},
  year={2024},
  note={DOI 10.52202/079017-2636}
}

@inproceedings{ruan2024toolemu,
  title={Identifying the Risks of {LM} Agents with an {LM}-Emulated Sandbox},
  author={Ruan, Yangjun and Dong, Honghua and Wang, Andrew and Pitis, Silviu and Zhou, Yongchao and Ba, Jimmy and Dubois, Yann and Maddison, Chris J. and Hashimoto, Tatsunori},
  booktitle={International Conference on Learning Representations},
  year={2024},
  note={DOI 10.48550/arXiv.2309.15817}
}

@article{debenedetti2025camel,
  title={Defeating Prompt Injections by Design},
  author={Debenedetti, Edoardo and Shumailov, Ilia and Fan, Tianqi and Hayes, Jamie and Carlini, Nicholas and Fabian, Daniel and Kern, Christoph and Shi, Chongyang and Terzis, Andreas and Tram{\`e}r, Florian},
  journal={arXiv preprint arXiv:2503.18813},
  year={2025},
  note={DOI 10.48550/arXiv.2503.18813}
}

@article{costa2025fides,
  title={Securing {AI} Agents with Information-Flow Control},
  author={Costa, Manuel and K{\"o}pf, Boris and Kolluri, Aashish and Paverd, Andrew and Russinovich, Mark and Salem, Ahmed and Tople, Shruti and Wutschitz, Lukas and Zanella-B{\'e}guelin, Santiago},
  journal={arXiv preprint arXiv:2505.23643},
  year={2025},
  note={DOI 10.48550/arXiv.2505.23643}
}

@article{li2026sure,
  title={``{Are You Sure?}'': An Empirical Study of Human Perception Vulnerability in {LLM}-Driven Agentic Systems},
  author={Li, Xinfeng and Dai, Shenyu and Zheng, Kelong and Xiao, Yue and Deng, Gelei and Dong, Wei and Wang, Xiaofeng},
  journal={arXiv preprint arXiv:2602.21127},
  year={2026},
  note={DOI 10.48550/arXiv.2602.21127}
}

@inproceedings{tang2026dark,
  title={Dark Patterns Meet {GUI} Agents: {LLM} Agent Susceptibility to Manipulative Interfaces and the Role of Human Oversight},
  author={Tang, Jingyu and Chen, Chaoran and Li, Jiawen and Zhang, Zhiping and Guo, Bingcan and Khalilov, Ibrahim and Gebreegziabher, Simret Araya and Yao, Bingsheng and Wang, Dakuo and Ye, Yanfang and Li, Tianshi and Xiao, Ziang and Yao, Yaxing and Li, Toby Jia-Jun},
  booktitle={Proceedings of the 2026 CHI Conference on Human Factors in Computing Systems},
  pages={403:1--403:26},
  year={2026},
  note={DOI 10.1145/3772318.3791568}
}

@article{grundemclaughlin2026overseeing,
  title={Overseeing Agents Without Constant Oversight: Challenges and Opportunities},
  author={Grunde-McLaughlin, Madeleine and Mozannar, Hussein and Murad, Maya and Chen, Jingya and Amershi, Saleema and Fourney, Adam},
  journal={arXiv preprint arXiv:2602.16844},
  year={2026},
  note={DOI 10.48550/arXiv.2602.16844}
}

@inproceedings{zhou2026check,
  title={When Should Users Check? {A} Decision-Theoretic Model of Confirmation Frequency in Multi-Step {AI} Agent Tasks},
  author={Zhou, Jieyu and Roy, Aryan and Gupta, Sneh and Weitekamp, Daniel and MacLellan, Christopher J.},
  booktitle={Proceedings of the 2026 CHI Conference on Human Factors in Computing Systems},
  year={2026},
  note={DOI 10.1145/3772318.3790655}
}

@article{michael2026permission,
  title={How Agents Ask for Permission: User Permissions for {AI} Agents, from Interfaces to Enforcement},
  author={Michael, Alexandra E. and Roesner, Franziska},
  journal={arXiv preprint arXiv:2607.13718},
  year={2026},
  note={DOI 10.48550/arXiv.2607.13718}
}

@article{wang2026laundering,
  title={Approval Laundering: Systematizing Approval-Execution Binding Failures in {AI} Coding-Agent Harnesses},
  author={Wang, Yang},
  journal={arXiv preprint arXiv:2609.38983},
  year={2026},
  note={DOI 10.48550/arXiv.2609.38983}
}

@inproceedings{amershi2019guidelines,
  title={Guidelines for Human-{AI} Interaction},
  author={Amershi, Saleema and Weld, Dan and Vorvoreanu, Mihaela and Fourney, Adam and Nushi, Besmira and Collisson, Penny and Suh, Jina and Iqbal, Shamsi and Bennett, Paul N. and Inkpen, Kori and Teevan, Jaime and Kikin-Gil, Ruth and Horvitz, Eric},
  booktitle={Proceedings of the 2019 CHI Conference on Human Factors in Computing Systems},
  year={2019},
  note={DOI 10.1145/3290605.3300233}
}

@incollection{hart1988nasa,
  title={Development of {NASA-TLX} ({Task Load Index}): Results of Empirical and Theoretical Research},
  author={Hart, Sandra G. and Staveland, Lowell E.},
  booktitle={Human Mental Workload},
  editor={Hancock, Peter A. and Meshkati, Najmedin},
  series={Advances in Psychology},
  volume={52},
  pages={139--183},
  publisher={North-Holland},
  year={1988},
  note={DOI 10.1016/S0166-4115(08)62386-9}
}

@article{parasuraman1997humans,
  title={Humans and Automation: Use, Misuse, Disuse, Abuse},
  author={Parasuraman, Raja and Riley, Victor},
  journal={Human Factors},
  volume={39},
  number={2},
  pages={230--253},
  year={1997},
  note={DOI 10.1518/001872097778543886}
}

@article{parasuraman2010complacency,
  title={Complacency and Bias in Human Use of Automation: An Attentional Integration},
  author={Parasuraman, Raja and Manzey, Dietrich H.},
  journal={Human Factors},
  volume={52},
  number={3},
  pages={381--410},
  year={2010},
  note={DOI 10.1177/0018720810376055}
}

@article{endsley1995toward,
  title={Toward a Theory of Situation Awareness in Dynamic Systems},
  author={Endsley, Mica R.},
  journal={Human Factors},
  volume={37},
  number={1},
  pages={32--64},
  year={1995},
  note={DOI 10.1518/001872095779049543}
}

@article{hutchins1985direct,
  title={Direct Manipulation Interfaces},
  author={Hutchins, Edwin L. and Hollan, James D. and Norman, Donald A.},
  journal={Human-Computer Interaction},
  volume={1},
  number={4},
  pages={311--338},
  year={1985},
  note={DOI 10.1207/s15327051hci0104\_2}
}

@inproceedings{bohme2010trained,
  title={Trained to Accept? {A} Field Experiment on Consent Dialogs},
  author={B{\"o}hme, Rainer and K{\"o}psell, Stefan},
  booktitle={Proceedings of the SIGCHI Conference on Human Factors in Computing Systems},
  pages={2403--2406},
  year={2010},
  note={DOI 10.1145/1753326.1753689}
}

@inproceedings{akhawe2013alice,
  title={Alice in Warningland: A Large-Scale Field Study of Browser Security Warning Effectiveness},
  author={Akhawe, Devdatta and Felt, Adrienne Porter},
  booktitle={Proceedings of the 22nd USENIX Security Symposium},
  pages={257--272},
  year={2013},
  note={\url{https://www.usenix.org/conference/usenixsecurity13/technical-sessions/paper/akhawe}}
}

@inproceedings{felt2012android,
  title={Android Permissions: User Attention, Comprehension, and Behavior},
  author={Felt, Adrienne Porter and Ha, Elizabeth and Egelman, Serge and Haney, Ariel and Chin, Erika and Wagner, David},
  booktitle={Proceedings of the Eighth Symposium on Usable Privacy and Security},
  year={2012},
  note={DOI 10.1145/2335356.2335360}
}

@inproceedings{bansal2021whole,
  title={Does the Whole Exceed Its Parts? {T}he Effect of {AI} Explanations on Complementary Team Performance},
  author={Bansal, Gagan and Wu, Tongshuang and Zhou, Joyce and Fok, Raymond and Nushi, Besmira and Kamar, Ece and Ribeiro, Marco Tulio and Weld, Daniel S.},
  booktitle={Proceedings of the 2021 CHI Conference on Human Factors in Computing Systems},
  year={2021},
  note={DOI 10.1145/3411764.3445717}
}

@article{bucinca2021trust,
  title={To Trust or to Think: Cognitive Forcing Functions Can Reduce Overreliance on {AI} in {AI}-Assisted Decision-Making},
  author={Bu{\c{c}}inca, Zana and Malaya, Maja Barbara and Gajos, Krzysztof Z.},
  journal={Proceedings of the ACM on Human-Computer Interaction},
  volume={5},
  number={CSCW1},
  pages={188:1--188:21},
  year={2021},
  note={DOI 10.1145/3449287}
}

@article{vasconcelos2023explanations,
  title={Explanations Can Reduce Overreliance on {AI} Systems During Decision-Making},
  author={Vasconcelos, Helena and J{\"o}rke, Matthew and Grunde-McLaughlin, Madeleine and Gerstenberg, Tobias and Bernstein, Michael S. and Krishna, Ranjay},
  journal={Proceedings of the ACM on Human-Computer Interaction},
  volume={7},
  number={CSCW1},
  pages={129:1--129:38},
  year={2023},
  note={DOI 10.1145/3579605}
}

@article{nissenbaum2004privacy,
  title={Privacy as Contextual Integrity},
  author={Nissenbaum, Helen},
  journal={Washington Law Review},
  volume={79},
  number={1},
  year={2004},
  note={\url{https://digitalcommons.law.uw.edu/wlr/vol79/iss1/10}}
}

@inproceedings{enck2010taintdroid,
  title={{TaintDroid}: An Information-Flow Tracking System for Realtime Privacy Monitoring on Smartphones},
  author={Enck, William and Gilbert, Peter and Chun, Byung-Gon and Cox, Landon P. and Jung, Jaeyeon and McDaniel, Patrick and Sheth, Anmol N.},
  booktitle={Proceedings of the 9th USENIX Symposium on Operating Systems Design and Implementation},
  pages={393--407},
  year={2010},
  note={\url{https://www.usenix.org/conference/osdi10/taintdroid-information-flow-tracking-system-realtime-privacy-monitoring}}
}

@article{llama32024,
  title={The {Llama} 3 Herd of Models},
  author={Grattafiori, Aaron and Dubey, Abhimanyu and Jauhri, Abhinav and Pandey, Abhinav and others},
  journal={arXiv preprint arXiv:2407.21783},
  year={2024},
  note={DOI 10.48550/arXiv.2407.21783}
}

@article{qwen32025,
  title={{Qwen3} Technical Report},
  author={Yang, An and Li, Anfeng and Yang, Baosong and Zhang, Beichen and others},
  journal={arXiv preprint arXiv:2505.09388},
  year={2025},
  note={DOI 10.48550/arXiv.2505.09388}
}
\end{document}